\documentclass[10pt,twocolumn]{article}

\usepackage[letterpaper,margin=0.72in]{geometry}
\usepackage{amsmath,amssymb}
\usepackage{booktabs}
\usepackage{float}
\usepackage{graphicx}
\usepackage{microtype}
\usepackage{iftex}
\ifPDFTeX
  \usepackage{times}
\else
  \usepackage{fontspec}
\fi
\usepackage[numbers,sort&compress]{natbib}
\usepackage[hidelinks]{hyperref}
\usepackage{url}
\usepackage{xspace}
\usepackage{balance}
\usepackage{placeins}

\newcommand{\method}{MemCorr-DP\xspace}

\newcommand{\romavtwo}{RoMa v2\xspace}
\newcommand{\visualunet}{Visual-DP-U-Net\xspace}
\newcommand{\visualtx}{Visual-DP-Transformer\xspace}

\title{\vspace{-0.65em}\method: Counterfactual Correspondence Conditioning\\
for a Diffusion Policy Guided by a Reference}
\author{%
Tan Su$^{1,*}$\quad Haoxiang Yang$^{2,*}$\quad
Ruxin Wang$^{1,*}$\quad Binghui Xie$^{3,\dagger}$\\[0.3em]
{\small $^{1}$Southern University of Science and Technology}\\
{\small $^{2}$Sun Yat-sen University\qquad $^{3}$Knowin}\\[0.2em]
{\footnotesize\texttt{12311316@mail.sustech.edu.cn}\quad
\texttt{yanghx29@mail2.sysu.edu.cn}}\\
{\footnotesize\texttt{12311507@mail.sustech.edu.cn}\quad
\texttt{xiebinghui0304@gmail.com}}\\[0.2em]
{\footnotesize $^{*}$Equal contribution (co-first authors).\qquad
$^{\dagger}$Corresponding author.}%
}
\date{}
\hypersetup{%
  pdftitle={MemCorr-DP: Counterfactual Correspondence Conditioning for a Diffusion Policy Guided by a Reference},
  pdfauthor={Tan Su, Haoxiang Yang, Ruxin Wang, Binghui Xie}%
}

\begin{document}
\maketitle
\vspace{-1.0em}

\begin{abstract}
Behavior-cloned visuomotor policies can remain accurate near their training distribution yet fail when object position and camera viewpoint change together. A successful reference trajectory contains the geometry needed to transfer the same interaction, but the policy must align that geometry with the current scene and remain sensitive to it during denoising. To address these challenges, we present \method, a diffusion policy that lifts frozen \romavtwo matches into explicit 3D relations between the current scene and the reference trajectory. A counterfactual paired objective assigns opposite behaviors the same physical state and noisy action while retaining reference-specific denoising targets. Mixed-condition fine-tuning then adapts the policy from ground-truth geometry to measured correspondence errors. Our strongest evaluation places the Door in the outermost position bands beyond the training support and changes the query camera by $\pm15^\circ$. Under this combined shift, \method achieves 96.67\% closed-loop success, compared with 88.00\% for a visual Transformer with the same action architecture. Objective ablations and reference interventions show that behavior responds to the selected reference, while matched controls favor the complete relation set over future motion or centroid geometry alone. These results support explicit 3D reference relations as a robust conditioning interface when spatial and viewpoint changes are compounded in the evaluated task.
\end{abstract}

\section{Introduction}

Behavior cloning can learn precise visuomotor skills from demonstrations, but prediction errors can move the policy away from the demonstrated state distribution~\citep{ross2011dagger}. The resulting policies can also depend strongly on the visual and spatial distribution seen during training. When an object is displaced or the camera viewpoint changes, the task remains the same while image structure and the relation between the robot and object change. A policy that relies on correlations tied to familiar positions or views can therefore fail even when the required interaction is unchanged. Robust generalization requires preserving the geometry of the demonstrated interaction and adapting it to the current scene.

A previously recorded successful trajectory of the target skill serves as the reference in this work. It contains the RGB-D observations and gripper motion from one successful execution, and therefore describes how the gripper moves relative to the object as the task progresses. Transferring this interaction to a new scene presents two coupled challenges. The policy must first establish a spatial correspondence between the recorded trajectory and the current observation. It must then retain the behavioral information carried by that correspondence while generating actions. The second challenge is especially important for diffusion policies because the noisy action can partially reveal the underlying behavior, allowing the denoiser to rely weakly on the reference. Visual matching errors further create a gap between geometric conditions available during training and those observed during execution.

To bridge this gap, we propose \method, a diffusion policy that expresses the reference trajectory through explicit 3D relations to the current scene. At every replanning step, frozen \romavtwo~\citep{edstedt2025romav2} matches the reference and current RGB-D observations, and valid matches are lifted into a shared 3D frame. The resulting relation set describes the live gripper around current scene geometry, the recorded gripper around reference geometry, future motion in the reference, and displacement between the two scenes. An action diffusion Transformer attends to these relations throughout denoising.

\begin{figure*}[t]
    \centering
    \includegraphics[width=0.98\textwidth]{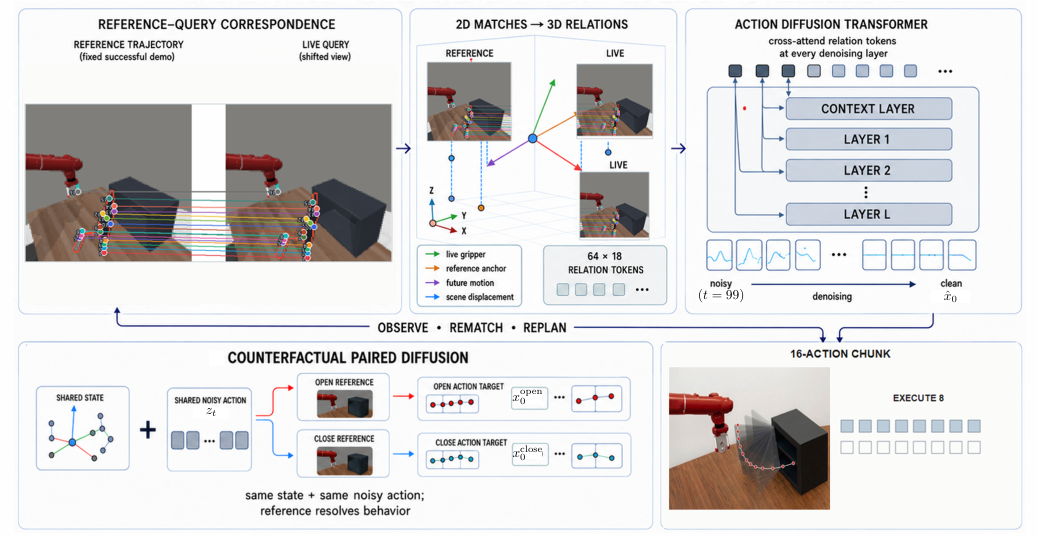}
    \caption{\textbf{End-to-end \method pipeline.} \emph{Top:} frozen \romavtwo matches a fixed successful reference to the live query. Up to sixteen sampled reference points are matched and lifted into a shared 3D frame. Four future offsets per point produce a fixed $64\times18$ relation-token array; invalid slots are masked. Green, orange, purple, and blue vectors denote the live gripper, reference anchor, future motion, and scene displacement. The Transformer uses these tokens throughout 20 DDIM steps. \emph{Bottom:} paired training shares state and noisy input $z_t$ across opening and closing. Distinct references and clean action targets $x_0^b$ define the behavior-specific denoising targets. The policy predicts 16 actions, executes eight, and then rematches and replans.}
    \label{fig:overview}
\end{figure*}

\begin{figure*}[t]
    \centering
    \includegraphics[width=0.88\textwidth]{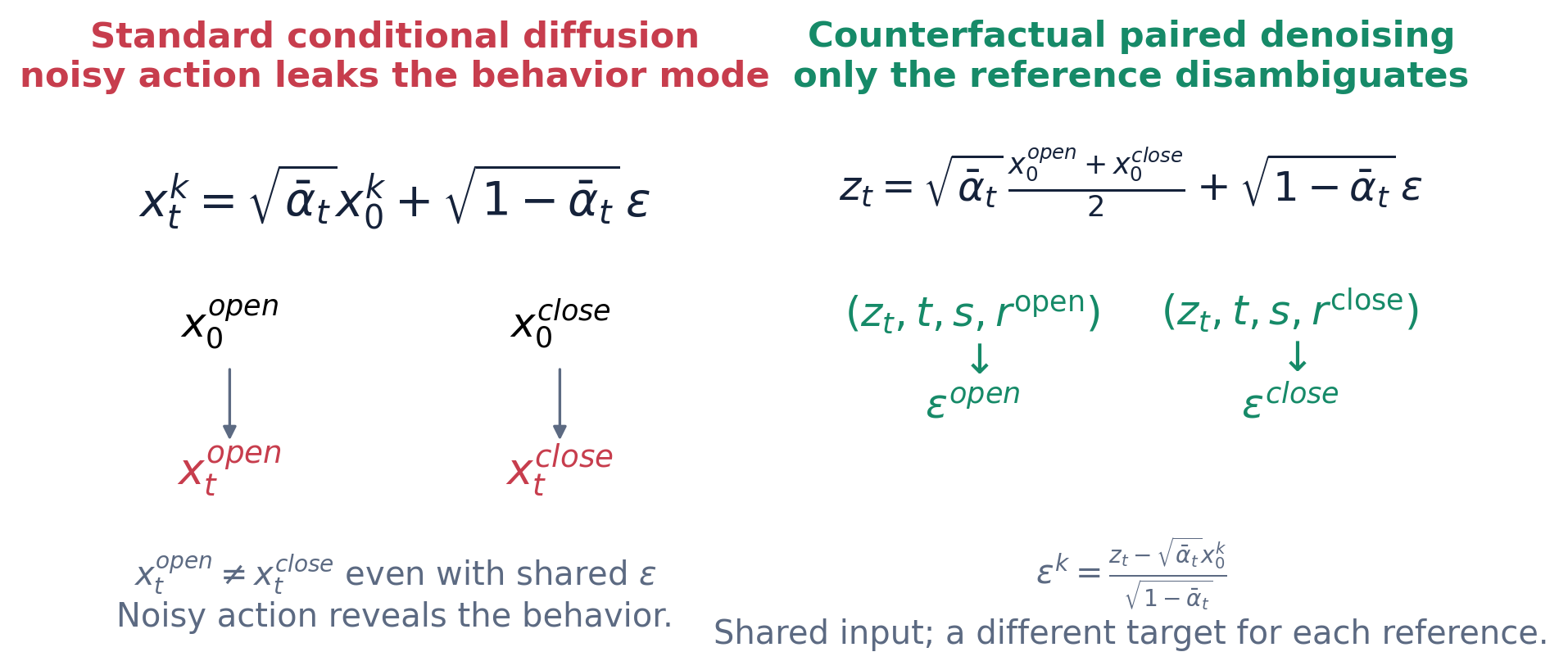}
    \caption{\textbf{Standard and counterfactual paired denoising.} In standard conditional diffusion, the clean action for each behavior produces a different noisy input. Counterfactual paired denoising uses the same robot state $s$ and noisy action $z_t$ for both behaviors while changing the reference and denoising target, as shown explicitly in the two right-hand input tuples.}
    \label{fig:paired}
\end{figure*}

To make behavior depend on the selected reference, we introduce a counterfactual paired objective. Opposite behaviors are trained with the same robot state and noisy action but with different references and denoising targets, removing the behavior cue from the shared noisy input. Fig.~\ref{fig:paired} contrasts these standard and counterfactual diffusion inputs. We then fine-tune with a mixture of ground-truth relations, measured RoMa v2 correspondences, and corruptions sampled from observed matching residuals. This two-stage procedure combines clean geometric supervision with adaptation to the correspondence errors encountered during execution.

The primary evaluation examines a combined position and viewpoint shift. Boundary extrapolation provides a complementary comparison. Objective ablations, reference interventions, and representation controls examine how correspondence conditioning affects the generated actions. Fig.~\ref{fig:overview} summarizes the complete pipeline, from reference--query matching and 3D relation construction to paired training and receding-horizon control.

Our contributions are:
\begin{itemize}
    \item a 3D relation representation for diffusion control that connects current scene geometry with demonstrated gripper state and future reference motion;
    \item a counterfactual paired denoising objective that gives opposite behaviors an identical state and noisy action, requiring the reference to resolve their distinct action targets;
    \item a two-stage training strategy that learns from exact geometric relations and then adapts to measured correspondence errors through mixed visual conditions and empirical residual corruption.
\end{itemize}

\section{Related Work}

\paragraph{Generalization in visuomotor diffusion policies.}
Diffusion Policy models multimodal action sequences through conditional denoising and receding horizon control~\citep{chi2023diffusion}. Although the standard formulation conditions on images, it does not specify how actions should transform when scene geometry changes. Subsequent methods target particular distribution shifts by adding representations or inductive biases. DP3 uses sparse point clouds, 3D Diffuser Actor embeds scene and noisy action tokens in a shared 3D space, and GenDP supplements 3D geometry with semantic fields tied to the task~\citep{ze2024dp3,ke2025diffuseractor,wang2025gendp}. Spatial mid-level experts and Render and Diffuse instead supply actionable perceptual cues or project candidate actions into image space~\citep{yang2025midlevel,vosylius2024render}. Equivariant Diffusion Policy and EquiBot impose $\mathrm{SO}(2)$ and $\mathrm{SIM}(3)$ transformation structure, while simpler symmetry methods use relative trajectories and symmetric visual encoders~\citep{wang2025equivariant,yang2024equibot,wang2025symmetry}. AffordDP transfers contact points and motion after contact as conditions and sampling guidance~\citep{wu2025afforddp}. Other work refines corrupted conditions or makes the diffusion source depend on the condition~\citep{chen2025noisyconditions,dong2025conditioning}. Across these methods, improvements are carried by explicit representations and priors rather than action diffusion alone; action memorization analyses further caution that success with limited demonstrations can resemble retrieval~\citep{he2026demystifying}.

\paragraph{Pixel correspondence for manipulation.}
Pixel correspondences link the same object regions across images and provide an explicit basis for geometric transfer. Dense Object Nets learn descriptors that are consistent across views, while LoFTR, DKM, RoMa, and RoMa v2 estimate dense matches across viewpoint and appearance changes~\citep{florence2018dense,sun2021loftr,edstedt2023dkm,edstedt2023roma,edstedt2025romav2}. In manipulation, Interaction Warping transfers a demonstrated interaction, R-NDF and TAX-Pose encode relative object placement, and Flow as the Cross-Domain Manipulation Interface transfers observed object motion~\citep{biza2023interaction,simeonov2023rndf,pan2023taxpose,xu2025flow}. Diffusion-PbD anchors an extracted waypoint program to matched image points and executes it with motion planning, whereas Knowledge-Driven Imitation Learning matches a semantic keypoint graph before passing its ordered points to a diffusion policy~\citep{murray2024diffusionpbd,miao2025knowledge}. CordViP and MatchingPolicy instead condition learned policies on correspondence features~\citep{fu2025cordvip,she2026matchingpolicy}. These formulations establish correspondence as a useful transfer interface. They leave open the joint use of 3D relations between live and reference points and future gripper motion from the selected reference at every replanning step of low-level action denoising.

\FloatBarrier
\section{Method}

\subsection{Observation and Reference Inputs}

At replanning step $k$, the policy receives a live observation
\begin{equation}
\mathcal O_k^q=(I_k^q,D_k^q,M_k^q,s_k^q,g_k^q),
\end{equation}
where $I_k^q\in\mathbb R^{H\times W\times3}$ and $D_k^q\in\mathbb R^{H\times W}$ are the RGB image and metric depth, $M_k^q\in\{0,1\}^{H\times W}$ is the visible object mask, $s_k^q\in\mathbb R^9$ is the robot state, and $g_k^q\in\mathbb R^3$ is the gripper position. The second input is a selected reference trajectory
\begin{equation}
\mathcal D^r=\{(I_j^r,D_j^r,M_j^r,s_j^r,g_j^r)\}_{j=0}^{T_r-1},
\end{equation}
where $I_j^r\in\mathbb R^{H\times W\times3}$, $D_j^r\in\mathbb R^{H\times W}$, $M_j^r\in\{0,1\}^{H\times W}$, $s_j^r\in\mathbb R^9$, and $g_j^r\in\mathbb R^3$ denote reference RGB, metric depth, visible object mask, robot state, and gripper position. The 9D state concatenates the end-effector and two fingertip positions, and all gripper positions are expressed in a common 3D world frame. The index $j_k$ denotes the current reference phase. Task identity is conveyed only by selecting the reference trajectory; the correspondence policy receives no separate task code. Section~\ref{sec:task_data} specifies the evaluated task, reference selection, and phase schedule.

\subsection{Object Centered Matching and 3D Lifting}

We run a frozen RoMa v2 matcher on object centered reference and query crops. Sixteen visible reference pixels are mapped to query locations with confidence $c_i\in[0,1]$, and the corresponding depth observations are lifted into a shared simulator world frame:
\begin{equation}
p_i^r=\Pi_r^{-1}(u_i^r,D^r(u_i^r)),\qquad
p_i^q=\Pi_q^{-1}(u_i^q,D^q(u_i^q)).
\end{equation}
Invalid lifts and matches outside the visible mask of the query object are excluded by a validity mask. Appendix~\ref{app:matching} specifies the RoMa v2 version, crop and point sampling, coordinate conversion, depth lifting, error diagnostic, and fallback behavior.

\subsection{Relational Condition Tokens}

For each point $i$ and slot $\ell\in\{0,1,2,3\}$ for a future offset, with $\tau_\ell\in\{0,5,10,15\}$, we construct
\begin{equation}
\begin{split}
r_{i,\ell}=\big[&g_k^q-p_i^q,\;g_{j_k}^r-p_i^r,\;
g_{j_k+\tau_\ell}^r-p_i^r,\\
&p_i^q-p_i^r,\;c_i,\;\bar\tau_\ell,\;\mathbf 0_4\big]\in\mathbb R^{18},
\end{split}
\label{eq:token}
\end{equation}
where $\bar\tau_\ell=\ell/3\in[0,1]$ indexes the sampled future offset and is distinct from the global reference phase $j_k$. The four 3D blocks respectively localize the live gripper around live geometry, anchor the demonstrated gripper at the current reference frame, describe its future motion around the same point, and encode displacement between the query and reference scenes. Sixteen points and four offsets produce 64 tokens; invalid tokens are masked. Field order, units, padding, normalization, and construction of the controls are specified in Appendix~\ref{app:architectures}.

\subsection{Action Diffusion with Relation Conditions}

The policy has 1.61M parameters. It encodes the 64 relation tokens and 9D robot state, then uses four Transformer action blocks~\citep{vaswani2017attention} to denoise $x_0\in\mathbb R^{16\times4}$. Each action contains a 3D end-effector displacement and one scalar gripper command. Each block applies self-attention over actions followed by cross-attention to the state and relation context. Training uses the standard noise-prediction objective of denoising diffusion~\citep{ho2020ddpm}; inference uses a deterministic DDIM sampler~\citep{song2021ddim} with 20 steps and classifier-free guidance~\citep{ho2022cfg}. Table~\ref{tab:architectures} gives the full network and Appendix~\ref{app:training} gives the diffusion schedule, normalization, and guidance implementation.

\subsection{Counterfactual Paired Denoising}

For behavior $b$, standard diffusion produces
\begin{equation}
x_t^b=\sqrt{\bar\alpha_t}x_0^b+\sqrt{1-\bar\alpha_t}\epsilon.
\end{equation}
Even if open and close share $\epsilon$, their different clean chunks make $x_t^{\mathrm{open}}\neq x_t^{\mathrm{close}}$; the noisy action leaks behavior identity. For paired chunks sharing the same physical initial condition, we instead define
\begin{align}
m&=\tfrac12(x_0^{\mathrm{open}}+x_0^{\mathrm{close}}),\\
z_t&=\sqrt{\bar\alpha_t}m+\sqrt{1-\bar\alpha_t}\epsilon,\\
\epsilon^b&=\frac{z_t-\sqrt{\bar\alpha_t}x_0^b}{\sqrt{1-\bar\alpha_t}}.
\end{align}
Both directions receive identical $(z_t,t,s)$ and differ in reference $r_b$ and target $\epsilon^b$:
\begin{equation}
\mathcal L_{\mathrm{pair}}=\frac12\sum_{b\in\{\mathrm{open},\mathrm{close}\}}
\operatorname{MSE}\!\left(\epsilon^b,\hat\epsilon_\theta(z_t,t,s,r_b)\right).
\label{eq:pairloss}
\end{equation}
Each pair uses the opening and closing chunks at frame zero from the same physical configuration. Training alternates standard and paired objectives at successive optimizer updates. Fig.~\ref{fig:overview} places the paired branch in the complete policy pipeline, while Fig.~\ref{fig:paired} isolates its contrast with standard conditional diffusion. Appendix~\ref{app:training} gives the sampling and batching details. The paired term is an auxiliary denoising objective rather than a replacement for the standard diffusion forward process. Standard updates retain the Gaussian noise prediction objective, and inference uses the same DDIM sampler.

\subsection{Mixture of Ground Truth and Visual Conditions}

Starting from the model trained with ground truth relations, we fine-tune for 10,000 updates with 25\% ground-truth, 50\% cached RoMa v2, and 25\% corrupted relation conditions in each batch. Corruptions resample measured matching residuals and validity patterns, with the same corruption shared by both directions of a counterfactual pair. Appendix~\ref{app:training} describes cache construction, mixture statistics, and corruption thresholds.

\section{Experimental Protocol}

\subsection{Task, Data, and Reference Protocol}
\label{sec:task_data}

We evaluate an open/close variant of a single articulated Door instance in Meta-World~\citep{yu2020metaworld}. Each paired physical configuration starts from the same Door joint value ($-0.7854$ rad), RGB-D observation, and robot state, but the two tasks require opposite actions. A physical configuration fixes the randomized Door $(x,y)$ position, camera, initial Door joint, and initial robot state. Door joint state, world object pose, target pose, and scene identity are excluded from policy inputs.

The dataset contains 100 paired physical configurations, each with one successful opening and closing trajectory of 200 steps. We use pairs 0--79 for training, yielding 160 trajectories and 29,600 action windows, and reserve pairs 80--99 for validation. All positions are sampled from the normalized range $[0.2,0.8]^2$.

During training, the trajectory from each physical configuration uses the trajectory from the next training configuration as its reference for the same skill, with the final configuration wrapping back to the first. For every query frame, we select the reference frame whose Door joint value is closest. During evaluation, opening and closing each use a fixed successful reference trajectory from training pair 0. The reference advances according to elapsed execution steps through Eq.~\eqref{eq:phase}, while the current rollout supplies the query observations.

Specifically, the reference phase is
\begin{equation}
j_k=\operatorname{clip}\!\left(\operatorname{round}(\rho e_k)+\delta,0,184\right),
\label{eq:phase}
\end{equation}
where $e_k$ is the number of executed environment steps, with default speed $\rho=1$ and offset $\delta=0$. The upper bound ensures $j_k+\max_\ell\tau_\ell\le199$. Each policy predicts 16 actions, executes the first eight, and then observes and replans.

\begin{figure*}[t]
    \centering
    \includegraphics[width=0.98\textwidth]{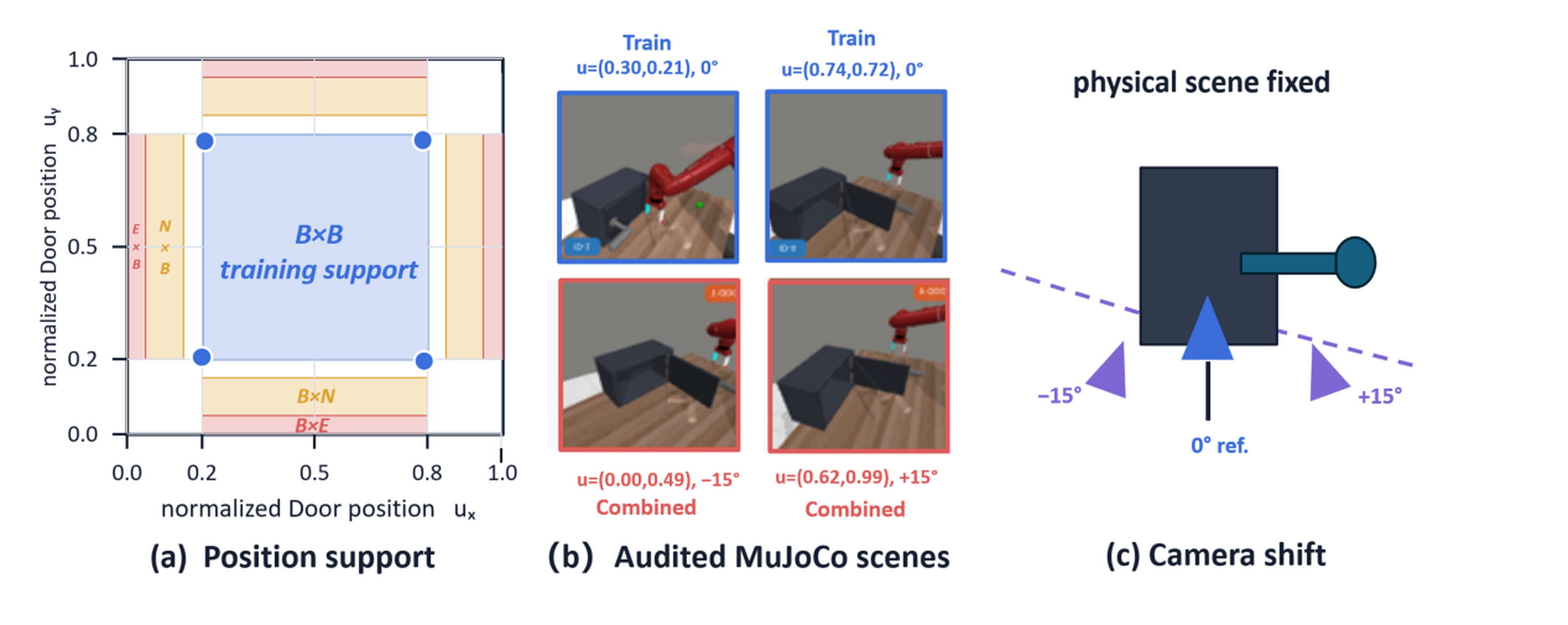}
    \caption{\textbf{Factorized position and viewpoint evaluation.} \textbf{(a)} Training uses $B\times B$, with $B=[0.2,0.8]$. Boundary and extreme tests place one coordinate in $N=[0.05,0.15]\cup[0.85,0.95]$ or $E=[0,0.05]\cup[0.95,1]$, respectively, while keeping the other in $B$. \textbf{(b)} Blue frames show representative training scenes; red frames show combined-shift scenes that pair an extreme position with a $\pm15^\circ$ query-camera offset. \textbf{(c)} Camera tests fix the physical scene and reference camera and rotate only the query camera.}
    \label{fig:ood}
\end{figure*}

All trainable policy components start from random initialization. Stage one trains for 20,000 updates on ground-truth relations and reports its final EMA. Stage two fine-tunes this EMA for 10,000 updates on mixed conditions and reports the resulting EMA. RoMa v2 remains frozen, and policy training uses seeds 42, 43, and 44. Appendix~\ref{app:training} gives the optimizer, learning rates, precision, batching, and checkpoint details. Fig.~\ref{fig:ood} summarizes the factorized position and query-view interventions.

\subsection{Baselines and Controls}

\paragraph{Visual Transformer with a shared denoiser.}
Our primary system baseline, \visualtx, replaces the relation tokenizer with current RGB, depth, visible object mask, and an explicit open/close code. It retains the \method action denoiser, data, optimization and diffusion schedules, horizons, seeds, and checkpoint rule. Its total trainable parameter count differs by $+0.36\%$. This experiment therefore compares two complete conditioning systems while holding the action denoiser, training budget, and rollout noise protocol fixed; the systems still differ in input information and perception.

\paragraph{Conventional U-Net visual policy.}
\visualunet provides a secondary conventional Diffusion Policy reference with comparable trainable capacity. It uses the same visual inputs, task code, data, horizon, update budget, diffusion schedule, seeds, and checkpoint rule as \visualtx. Its 1D U-Net denoiser uses the standard objective on independent samples without condition dropout, so this comparison also changes the architecture and objective.

Both visual policies receive a 2D one-hot task code directly, whereas \method uses the known task to select a fixed reference for the same skill. Appendix~\ref{app:architectures} specifies all three networks and their trainable parameter counts.

\paragraph{Matched representation controls.}
\emph{Motion only} removes point geometry and repeats the future reference displacement $g^r_{j_k+\tau_\ell}-g^r_{j_k}$, confidence 1, and phase across the 16 point slots. \emph{Object centroid} replaces all valid reference and query points by their respective centroids before constructing Eq.~\eqref{eq:token}; invalid slots remain masked. Both preserve the $64\times18$ interface, Transformer, training objective, budget, horizons, and seeds.

\subsection{Fixed Position and Viewpoint Tests}

Door position varies in two dimensions. Simulator ranges $x\in[0,0.10]$ m and $y\in[0.85,0.95]$ m are mapped independently to normalized coordinates $u\in[0,1]^2$; Door height remains fixed. Table~\ref{tab:test_ranges} summarizes all position and camera distributions. Training and validation use Latin hypercube sampling, whereas test positions are sampled uniformly. For the boundary and extreme position tests, the active coordinate alternates between the two axes and between the lower and upper bands. Camera interventions affect only the query view; the reference camera remains fixed.

\begin{table}[t]
\centering
\caption{Training and test distributions. $B=[0.2,0.8]$ is the training band, $N=[0.05,0.15]\cup[0.85,0.95]$ is the boundary band, and $E=[0,0.05]\cup[0.95,1]$ is the extreme band. The final column counts distinct physical scene setups; every test scene is evaluated for both tasks with three training seeds.}
\label{tab:test_ranges}
\footnotesize
\setlength{\tabcolsep}{3pt}
\begin{tabular}{lccc}
\toprule
Condition & Position $(u_x,u_y)$ & Query camera & Scenes \\
\midrule
Training / validation & $(B,B)$ & original & 80 / 20 \\
Boundary extrapolation & $(N,B)$ or $(B,N)$ & original & 50 \\
Extreme position & $(E,B)$ or $(B,E)$ & original & 25 \\
Camera shift & $(B,B)$ & $\pm15^\circ$ & 25 \\
Combined shift & $(E,B)$ or $(B,E)$ & $\pm15^\circ$ & 25 \\
\bottomrule
\end{tabular}
\end{table}

A scripted expert solved both directions within 200 steps in all 125 evaluation configurations, confirming that the tasks were feasible. Every method is evaluated on the same configurations.

\begin{figure*}[t]
    \centering
    \includegraphics[width=0.98\textwidth]{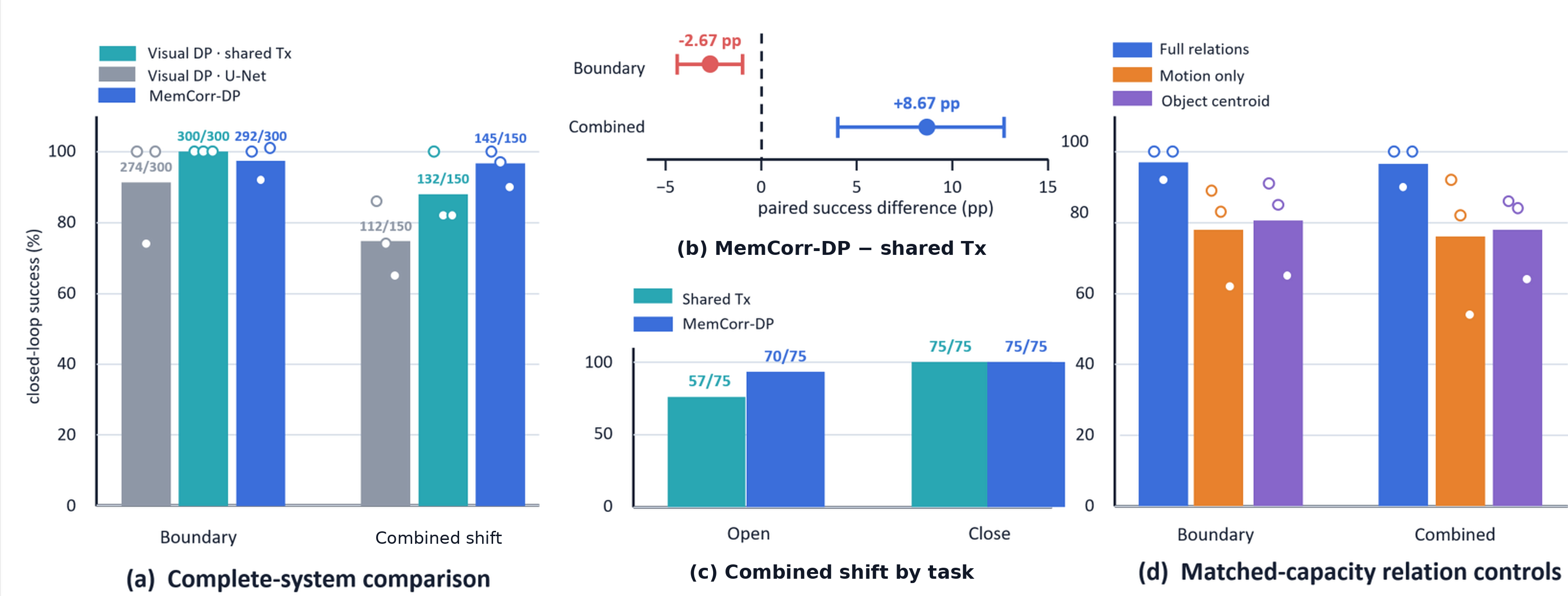}
    \caption{\textbf{Closed-loop system comparison and matched relation controls.} \textbf{(a)} Complete-system success under boundary extrapolation ($n=300$ per method) and the combined shift ($n=150$); bars pool three training seeds, circles show individual seeds, and labels give successes/total. \textbf{(b)} Paired difference between \method and \visualtx with 95\% physical-configuration bootstrap intervals; positive values favor \method. \textbf{(c)} Combined-shift success by task ($n=75$ per task and method). \textbf{(d)} Full relations versus separately trained motion-only and object-centroid controls with the matched Transformer architecture; marks follow panel (a).}
    \label{fig:main}
\end{figure*}

\begin{table*}[t]
\centering
\caption{\textbf{System and representation comparisons.} Success is reported as count/total (percentage). Parameters include all trainable policy modules and exclude frozen RoMa v2; motion and centroid controls are trained separately.}
\label{tab:system}
\small
\textbf{(a) Complete policy systems}\\[2pt]
\setlength{\tabcolsep}{6pt}
\begin{tabular}{lrrr}
\toprule
Method & Params & Boundary extrapolation & Combined shift \\
\midrule
Visual U-Net & 1.556M & 274/300 (91.33\%) & 112/150 (74.67\%) \\
Visual Transformer & 1.619M & \textbf{300/300 (100.00\%)} & 132/150 (88.00\%) \\
\method & 1.613M & 292/300 (97.33\%) & \textbf{145/150 (96.67\%)} \\
\bottomrule
\end{tabular}
\\[6pt]
\textbf{(b) Representation controls with matched architecture}\\[2pt]
\begin{tabular}{lrr}
\toprule
Condition & Boundary extrapolation & Combined shift \\
\midrule
Full relation condition & \textbf{292/300 (97.33\%)} & \textbf{145/150 (96.67\%)} \\
Motion only & 234/300 (78.00\%) & 114/150 (76.00\%) \\
Object centroid & 241/300 (80.33\%) & 117/150 (78.00\%) \\
\bottomrule
\end{tabular}
\end{table*}

\subsection{Metrics and Statistical Unit}

The primary metric is the official Meta-World success flag within 200 calls to the environment step function; each executed action counts as one step. Opening succeeds when the handle's $x$ coordinate is within 0.08 m of its target, and closing succeeds when the handle is within 0.08 m Euclidean distance of its target. No additional requirement is imposed on contact, gripper state, or holding success for consecutive steps. A rollout stops at first success; timeout or environment termination without success is a failure.

Boundary extrapolation is evaluated on 50 physical configurations, and each stronger condition uses 25. Every configuration is tested in both task directions with three training seeds, yielding 300 and 150 rollouts per condition, respectively. Matched comparisons use the same configuration, task, and training seed identifiers. For primary 95\% intervals, we first average tasks and seeds within each physical configuration and then resample configurations; paired differences are computed before aggregation. Appendix~\ref{app:evaluation} specifies rollout randomness and the bootstrap procedure. Fig.~\ref{fig:main} summarizes the complete-system comparison, task split, and matched representation controls.

\section{Results}

\subsection{Main System Comparison}

Table~\ref{tab:system} and Fig.~\ref{fig:main}(a,b) compare the complete systems. Relative to the visual Transformer, \method differs by $-2.67$ percentage points on boundary extrapolation, with a 95\% configuration bootstrap interval of $[-4.33,-1.00]$, and by $+8.67$ points under the combined shift, with an interval of $[4.00,12.67]$. Fig.~\ref{fig:main}(c) localizes the combined-shift difference to opening: \method solves 70/75 opening rollouts, compared with 57/75 for the visual Transformer, while both methods solve all 75 closing rollouts. The conventional \visualunet is lower in both conditions.

\subsection{Spatial Relations Exceed Simpler Reference Information}

Table~\ref{tab:system}(b) and Fig.~\ref{fig:main}(d) show that, with architecture and training schedule fixed, full relations outperform motion and centroid conditions in both tests. Every paired 95\% interval is above zero. This experiment compares complete representations rather than individual token fields.

The motion condition favors closing, whereas the centroid condition is stronger on opening and weaker on closing. Under the combined shift, we kept both point sets unchanged but reassigned query points to reference points and recomputed the affected relation fields. The same final checkpoints with ground-truth geometry achieved 130/150 successes with correct pairing and 148/150 after permutation. Correct point pairing was therefore not necessary for high success under this fixed permutation.

\subsection{Paired Denoising Training Reduces Direction Bypass}

\begin{figure}[t]
    \centering
    \includegraphics[width=\linewidth]{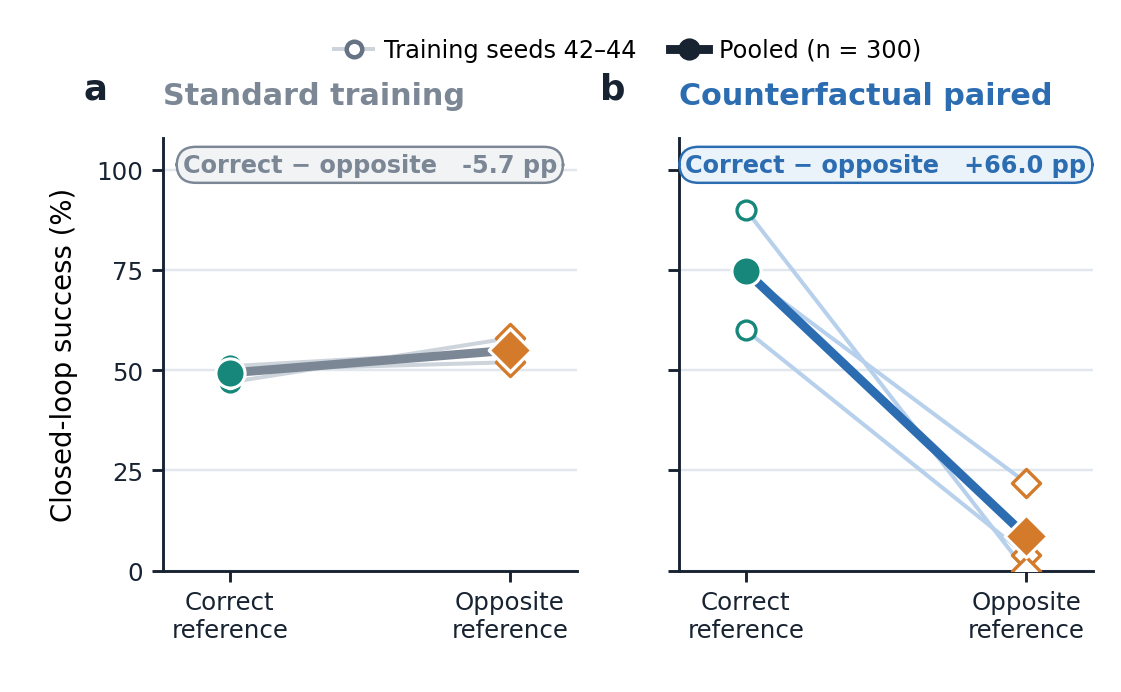}
    \caption{\textbf{Counterfactual paired training restores reference selectivity.} \textbf{(a,b)} Stage-one closed-loop success with ground-truth relations from correct and opposite references after standard and paired training. Success is scored against the original task; lower opposite-reference success therefore indicates suppression of original-task behavior, not completion of the supplied opposite task. Each thin line connects the two reference conditions for one training seed (42--44; $n=100$ per seed and condition). Thick lines pool all 300 boundary-extrapolation rollouts per condition. Circles and diamonds denote correct and opposite references; badges report the pooled correct-minus-opposite difference.}
    \label{fig:objective}
\end{figure}

Fig.~\ref{fig:objective} reports the stage-one reference intervention used to test whether the denoiser follows the selected behavior.

\begin{table}[t]
\centering
\caption{\textbf{Training ablation and reference intervention for boundary extrapolation.} Success is scored against the original task. Panel (a) uses stage-one checkpoints with ground-truth relations or an empty condition; panel (b) uses final checkpoints trained with mixed conditions.}
\label{tab:objective}
\footnotesize
\setlength{\tabcolsep}{3pt}
\textbf{(a) Training procedure}\\[2pt]
\begin{tabular}{llr}
\toprule
Training & Reference & Success \\
\midrule
Standard & Correct & 148/300 (49.33\%) \\
& Empty & 76/300 (25.33\%) \\
& Opposite & 165/300 (55.00\%) \\
\cmidrule(lr){1-3}
Counterfactual & Correct & \textbf{224/300 (74.67\%)} \\
paired & Empty & 118/300 (39.33\%) \\
& Opposite & \textbf{26/300 (8.67\%)} \\
\bottomrule
\end{tabular}
\\[5pt]
\textbf{(b) Reference condition}\\[2pt]
\begin{tabular}{lr}
\toprule
Condition & Success \\
\midrule
Ground-truth relation, correct & 277/300 (92.33\%) \\
\romavtwo, correct & \textbf{292/300 (97.33\%)} \\
Empty relation & 95/300 (31.67\%) \\
\romavtwo, opposite & 5/300 (1.67\%) \\
\bottomrule
\end{tabular}
\end{table}

Table~\ref{tab:objective}(a) and Fig.~\ref{fig:objective} show that, with architecture and training budget fixed, standard conditional diffusion retains success on the original task more often under an opposite reference than under the correct one. The counterfactual paired training procedure reverses this ordering across all three seeds: it improves success with the correct reference and suppresses the original task under an opposite reference. The first action chunk follows the supplied reference direction in every rollout of the paired model. This ablation evaluates the complete paired training procedure, including its paired sampling and high noise timestep range.

\subsection{Behavior Responds to Reference Content}

Table~\ref{tab:objective}(b) uses correct, empty, and opposite conditions to distinguish sensitivity to reference content from sensitivity to the presence of any condition. Under an opposite reference, success on the original task nearly vanishes, while the first action chunk follows the supplied direction in every rollout and final Door displacement follows it in 292/300. Because success remains defined by the original task, these directional measures indicate reference following rather than completion of the supplied task. Empty conditioning provides a control for dependence on the relation input.

The mixed checkpoint performs better with \romavtwo than with ground-truth correspondence. Because fine-tuning with mixed conditions is weighted toward raw or corrupted visual conditions, this difference is consistent with calibration to the distribution of visual conditions.

\subsection{Position, Viewpoint, and Matching Robustness}

\begin{table*}[t]
\centering
\caption{\textbf{Success under stronger shifts.} Each entry reports count/150 (percentage). Full RoMa v2 and ground-truth correspondence use the same final checkpoints; motion and centroid conditions are trained separately. Bold marks the highest success in each column; an em dash denotes an unmeasured condition.}
\vspace{2mm}
\label{tab:strong}
\small
\setlength{\tabcolsep}{7pt}
\begin{tabular}{lrrr}
\toprule
Condition & Extreme position & Camera shift & Combined shift \\
\midrule
Full \romavtwo & \textbf{145/150 (96.67\%)} & 148/150 (98.67\%) & \textbf{145/150 (96.67\%)} \\
Ground-truth correspondence & 130/150 (86.67\%) & \textbf{150/150 (100.00\%)} & 130/150 (86.67\%) \\
Motion only & 113/150 (75.33\%) & 119/150 (79.33\%) & 114/150 (76.00\%) \\
Object centroid & -- & -- & 117/150 (78.00\%) \\
\bottomrule
\end{tabular}
\end{table*}

Full \romavtwo attains similar success under the extreme position, camera, and combined shifts (Table~\ref{tab:strong}). The camera test changes only the query view within each physical configuration and therefore probes matching and 3D lifting without adding position extrapolation. The combined condition also requires transfer to extreme positions.

The mean 3D correspondence error increases from 11.78 mm for boundary extrapolation to 43.66 mm under the combined shift. Fallback due to an invalid condition affects 14 of 3,399 and 39 of 1,731 replanning decisions, respectively. The policy achieves high task success despite the larger matching error. For this diagnostic, simulator ground truth is used after rollout collection and is not provided during policy execution.

\subsection{Sensitivity to Reference Phase}

\begin{figure}[t]
    \centering
    \includegraphics[width=\linewidth]{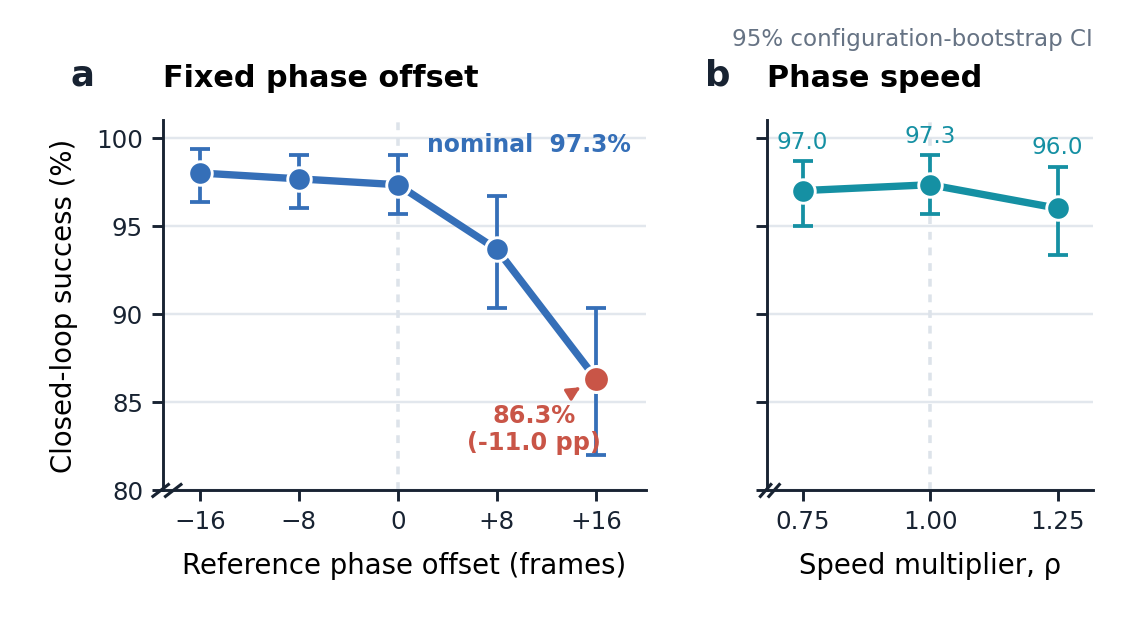}
    \caption{\textbf{Sensitivity to reference phase under boundary extrapolation.} \textbf{(a)} Fixed phase offsets; negative offsets lag the reference, whereas positive offsets lead it, and $+16$ frames is highlighted. \textbf{(b)} Phase-speed multipliers. Points pool 300 rollouts per condition; error bars are 95\% physical-configuration cluster-bootstrap intervals. Axes begin at 80\%.}
    \label{fig:clock}
\end{figure}

Changing the reference phase based on elapsed steps in Eq.~\eqref{eq:phase} shows tolerance to moderate offset and speed mismatch, while a reference lead of 16 frames reduces success (Fig.~\ref{fig:clock}).

\section{Discussion}

The central advantage of \method appears when object position and camera viewpoint change together. Under this combined shift, it achieves 96.67\% success versus 88.00\% for the visual Transformer with the same action denoiser. By placing live geometry, gripper state, and demonstrated future motion in a shared 3D frame, \method ties the control condition to interaction geometry rather than the pixel layout seen during training.

Matched controls and interventions support this interpretation: full relations outperform motion and centroid controls, counterfactual paired training makes actions respond to the selected reference, and high success persists as matching error grows. The point permutation diagnostic shows that this advantage cannot be attributed simply to retaining correct point pairing. Although the complete systems use different conditioning modalities, the matched controls isolate the representation more closely. Within the evaluated Door task, explicit 3D relations therefore provide a more robust interface under the stronger distribution shift.

\section{Limitations and Broader Impact}

The evaluation covers one simulated Door instance, two directions, translational position shifts, and camera azimuth changes of $\pm15^\circ$. It assumes known task identity, one fixed reference per task, simulator-provided object masks, calibrated RGB-D lifting, and an elapsed-time reference phase. The conclusions are therefore limited to this instance and these shifts; transfer across instances, categories, rotations, sensing domains, and physical robots remains untested.

Parameter matching excludes the frozen RoMa v2 backbone, and the two systems differ in observation bandwidth and inference cost. Three training seeds provide limited evidence about optimization variability. Physical deployment would additionally require robustness to segmentation, depth, calibration, contact, and actuator errors, as well as a validated safety mechanism.

\section{Conclusion}

\method conditions action diffusion on 3D relations between a reference trajectory and the live scene. Counterfactual paired training makes action direction responsive to the selected reference, and training with mixed conditions accommodates measured matching error. On the evaluated Door instance, the method achieves high success when position and viewpoint shift together and outperforms the visual policy with a matched action architecture in this stronger setting.

\FloatBarrier
\global\vfuzz=2pt
{\footnotesize
\setlength{\bibsep}{0pt plus 0.2pt}
\bibliographystyle{plainnat}
\bibliography{references}
}

\appendix

\section{Correspondence, Lifting, and Fallback}
\label{app:matching}

\paragraph{Matcher and pixel coordinates.}
The matcher is RoMa v2 release v2.0.1~\citep{edstedt2025romav2} with the ``turbo'' inference setting and a frozen DINOv3 ViT-L/16 descriptor~\citep{simeoni2025dinov3} using layers 11 and 17. Each original RGB, depth, and mask frame is $128\times128$. The tight bounding box of the nonzero mask is padded by 12 pixels on all sides and clipped to $[0,127]^2$ before cropping. Each RGB crop is resized to $128\times128$ for matching. Visible reference pixels are enumerated in row-major order; 16 indices are chosen by evenly spaced positions over this list. If fewer than 16 pixels exist, the last location is repeated to preserve shape and repeated slots are invalid. Selection is deterministic, and match confidence is used as a token value rather than a selection criterion.

RoMa warp and confidence fields are bilinearly sampled at the selected reference pixels using border padding. A noninteger mapped location is converted from crop to original coordinates using pixel center scaling. Query depth uses bilinear sampling, while visible-mask membership uses the nearest rounded pixel. Confidence is clipped to $[0,1]$ without thresholding.

\paragraph{Metric lifting and validity.}
Rendered depth buffers are converted to meters. A depth value is invalid if the coordinate is out of bounds, the value is nonfinite, or depth is nonpositive. Intrinsics are derived from the current MuJoCo camera field of view; extrinsics come from its current pose and are refreshed after every camera intervention or reset. Back projection first produces OpenCV camera coordinates and then applies the transform from camera to world coordinates. Thus $p_i^r$ and $p_i^q$ are both in the same simulator world frame even when reference and query cameras differ.

A slot is invalid if reference selection is padded, either lift is invalid or nonfinite, the warp is nonfinite or out of range, or the nearest query pixel lies outside the visible object mask. Validity uses the rendered mask without forward-backward consistency or a separate occlusion classifier. Invalid coordinates receive finite internal placeholders, while their 18D tokens are zeroed and excluded by the attention mask.

\paragraph{Ground truth and error diagnostic.}
Training and interventions with ground-truth relations transform 16 fixed 3D points in local handle coordinates by the Door pose in the reference and query scenes. For the full RoMa v2 diagnostic, each estimated reference point is mapped into the local reference Door frame, transported through the query Door transform, and compared with the estimated query point by Euclidean distance in world coordinates. The online mean first averages valid points within each replanning decision and then pools decisions across the three seed logs. Cache median, 95th, and 99th percentiles instead pool all 149,063 valid raw points; they are 3.72, 20.03, and 79.62 mm, respectively (mean 7.95 mm). Simulator ground truth for this diagnostic is computed after rollout collection and is not an online policy input.

\paragraph{Fallback for invalid conditions.}
If fewer than four valid raw points remain, all 64 relation tokens and their validity mask are set to zero/false. The policy continues sampling and executing actions from robot state and diffusion noise, without reusing previous relations or substituting a static or safety action. Validity is recomputed at every replan; no fallback state persists. Fallback rollouts remain in all success denominators.

\begin{table*}[t]
\centering
\caption{Trainable policy architectures. ``Pre-LN'' denotes layer normalization before the named block. Parameter totals include all listed modules and exclude frozen RoMa v2.}
\label{tab:architectures}
\small
\setlength{\tabcolsep}{4pt}
\begin{tabular}{p{0.17\textwidth}p{0.20\textwidth}p{0.55\textwidth}}
\toprule
Model & Component & Exact structure \\
\midrule
\method & Relation/context encoder & Linear $18\!\to\!128$; learned embeddings for 16 points and 4 offsets; 2 pre-LN Transformer encoder layers, width 128, 4 heads, MLP $128\!\to\!512\!\to\!128$, GELU, dropout 0. \\
& State/time/action inputs & State MLP $9\!\to\!128\!\to\!128$ with GELU; action Linear $4\!\to\!128$ plus 16 learned positional embeddings for actions; sinusoidal time embedding 128 followed by $128\!\to\!512\!\to\!128$ GELU MLP. \\
& Action denoiser/output & 4 pre-LN blocks, each self-attention (4 heads), cross-attention to one state token plus 64 context tokens, and GELU MLP $128\!\to\!512\!\to\!128$; dropout 0; final LayerNorm and Linear $128\!\to\!4$. Total 1,612,804. \\
\midrule
\visualtx & Image/task tokenizer & RGB, depth, and mask form $5\times64\times64$. One Conv2d $5\!\to\!18$, kernel/stride 8, yields $8\times8=64$ tokens. A Linear $2\!\to\!18$ one-hot task embedding is added to every token. Tokenizer total 5,832; no pretrained or frozen visual encoder. \\
& Shared action core & The complete \method core above with 1,612,804 parameters, including its $18\!\to\!128$ projection, context encoder, state encoder, time encoder, and action blocks. Total 1,618,636, or $+0.36\%$; the visual encoder is included. \\
\midrule
\visualunet & Observation encoders & Five stride-2 Conv2d--GroupNorm(4)--Mish blocks: $5\!\to\!24$ with kernel 5, then $24\!\to\!32\!\to\!48\!\to\!64\!\to\!96$ with kernel 3; global average pooling and $96\!\to\!96$ Linear--LayerNorm--Mish. State: $9\!\to\!32\!\to\!24$; task: $2\!\to\!8$. Concatenated global condition is 128D. \\
& Conditional 1D U-Net & Sinusoidal time dimension 64 with $64\!\to\!256\!\to\!64$ Mish MLP. Down widths 48, 96, 144; two conditional residual blocks per level, kernel 3 and GroupNorm(4), stride-2 downsampling. Two 144D middle blocks, symmetric up blocks with skip connections and kernel-4 stride-2 transposed convolution, and a final kernel-3 block plus $1\times1$ output. Each residual block obtains scale/bias from a Mish--Linear projection of time plus global condition. Total 1,556,228. \\
\bottomrule
\end{tabular}
\end{table*}

\section{Token Controls and Network Architectures}
\label{app:architectures}

\paragraph{Token interface.}
The fixed field order is
\[
\begin{aligned}\relax
[&\underbrace{g_k^q-p_i^q}_{1:3},\underbrace{g_{j_k}^r-p_i^r}_{4:6},
\underbrace{g_{j_k+\tau_\ell}^r-p_i^r}_{7:9},\\
 &\underbrace{p_i^q-p_i^r}_{10:12},\underbrace{c_i}_{13},
\underbrace{\ell/3}_{14},\underbrace{0,0,0,0}_{15:18}].
\end{aligned}
\]
All 3D values are measured in meters in the world coordinate frame; confidence and phase lie in $[0,1]$. Token fields are not normalized separately. The motion condition sets fields 1--12 to zero except that fields 7--9 contain $g^r_{j_k+\tau_\ell}-g^r_{j_k}$, sets confidence to one, and repeats the four offset tokens over 16 nominal point slots. The centroid control computes separate arithmetic centroids over valid reference and query points and substitutes those centroids in every valid point slot; invalid slots remain masked. The four padding fields are fixed to zero in every method. Attention masks exclude invalid slots after linear projection, including any projection bias.

\paragraph{Visual preprocessing.}
For both visual policies, rendered $128\times128$ RGB, depth, and mask observations are resized to $64\times64$ and concatenated as five channels. RGB is mapped from $[0,255]$ to $[-1,1]$. Depth is standardized using mean and standard deviation computed over all 32,000 frames of training pairs 0--79 and clipped to $[-5,5]$; the mask remains binary. There is no image augmentation. Both visual policies use the same standardized 9D robot state and 2D one-hot open/close code. The \visualtx tokenizer contributes 5,832 parameters, all included in its reported total.

\section{Data, Optimization, and Diffusion Details}
\label{app:training}

\paragraph{Window construction and normalization.}
Every trajectory of 200 steps contributes query anchors 0--184 so that the action window of 16 steps and the largest future offset 15 remain in bounds. Across 80 physical pairs and two directions this yields 29,600 windows. Standard updates uniformly sample row indices with replacement. Actions and 9D state are standardized with means and standard deviations from these training windows; state standard deviations are floored at $10^{-3}$ and action standard deviations at $5\times10^{-2}$. Clean targets, Gaussian noise, and denoiser predictions therefore share normalized action coordinates. Predicted clean actions are clipped to $[-5,5]$ in normalized space during inference, converted back to the original action units, and then clipped componentwise to $[-1,1]$.

\paragraph{Diffusion implementation.}
The forward schedule is $\beta_t=10^{-4}+t(2\times10^{-2}-10^{-4})/99$ and $\bar\alpha_t=\prod_{q=0}^{t}(1-\beta_q)$, so $t=0$ and $t=99$ are the lowest and highest noise levels. Standard updates independently sample $t\sim\mathcal U\{0,\ldots,99\}$ per example and IID standard Gaussian noise. Inference starts from an IID standard Gaussian $16\times4$ chunk and uses the descending integer indices from \texttt{linspace(0,99,20)}. The sampler is deterministic after its initial noise (DDIM stochasticity zero). Guidance drops only the condition of 64 tokens and uses $\hat\epsilon=\hat\epsilon_{u}+1.5(\hat\epsilon_{c}-\hat\epsilon_{u})$; robot state is present in both predictions, and \method has no task code. \visualtx applies the same rule to all visual tokens, thereby also removing its embedded task code in the unconditional branch. \visualunet does not use classifier-free guidance.

\paragraph{Counterfactual batches.}
A paired batch samples 64 training-pair indices with replacement and takes the opening and closing rows at frame 0 for each. The implementation verifies equality of the paired 9D states. The shared midpoint and Gaussian noise are constructed in normalized action space. One timestep is sampled per physical pair and repeated for its two directions. The loss is averaged over the 128 direction samples and every horizon and action coordinate. Paired updates omit condition dropout. Timesteps are sampled from $\{80,\ldots,99\}$ to concentrate this auxiliary supervision at high noise levels, where the clean action contributes less information to the noisy input.

\paragraph{Mixed cache and corruption.}
The cache has 10,000 rows and is generated once with seed 43001. It contains all 160 paired rows at frame 0 and 9,840 distinct noninitial training rows sampled without replacement. Pairing between reference and query remains cyclic across different trajectories for the same skill. Within every standard batch or batch of physical pairs, integer source counts realize exactly 25\% ground truth, 50\% raw RoMa cache, and 25\% residual bootstrap.

Residual-bootstrap corruption reuses the measured point-validity mask from the selected cache row. For each valid point, a 3D residual is sampled independently from the empirical pool after discarding vectors above 20.034 mm and added to the ground-truth query point. Reference points, masks, and confidence values come from the cache row, with confidence unchanged. A counterfactual pair shares the 16 sampled residuals, reference and query geometry, masks, and confidences; future reference gripper motion remains task specific.

\paragraph{Optimization.}
All models use batch size 128 and AdamW with $(\beta_1,\beta_2)=(0.9,0.999)$, $\epsilon=10^{-8}$, weight decay $10^{-4}$, no warm-up, and full precision training. The learning rate is $3\times10^{-4}$ for 20,000 updates and $5\times10^{-5}$ for the final 10,000 updates. Gradients are clipped to global norm 10, and EMA with decay 0.999 is updated after every optimizer step. \method and \visualtx alternate standard and paired updates in both stages, with probability 0.1 of dropping the condition only on standard updates. \visualunet uses standard updates throughout and no condition dropout. All methods share training pairs, horizons, update budget, and seeds. Ablations after stage one use the EMA after update 20,000; final policies use the EMA after update 30,000.

\balance
\section{Evaluation Statistics}
\label{app:evaluation}

\paragraph{Rollout randomness.}
Each method is evaluated once for every combination of condition, configuration, task, and training seed. Matched comparisons use the same initial configuration and generate action noise deterministically from the training seed, configuration, task, and replanning index.

\paragraph{Confidence intervals.}
Each physical configuration contributes the mean over its two task outcomes and three training seeds. We resample these configuration means 100,000 times with replacement and take the 2.5th/97.5th percentiles. For paired comparisons, we subtract outcomes within matched rollouts before the same averaging and resampling. These intervals quantify variation across tested configurations conditional on the three trained models, not uncertainty over repeated training studies.

\end{document}